\documentclass[11pt]{article}

\usepackage{cite}
\usepackage[utf8]{inputenc}
\usepackage[T1]{fontenc}
\usepackage{amsmath,amssymb,amsfonts}
\usepackage{algorithmic}
\usepackage{graphicx}
\usepackage{caption}
\usepackage{hyperref}
\usepackage{textcomp}
\usepackage{xcolor}
\usepackage{comment}
\usepackage[
    top=2.5cm,
    bottom=2.5cm,
    left=2.5cm,
    right=2.5cm
]{geometry}
\usepackage{booktabs}
\usepackage{tabularx}
\usepackage{authblk}
\usepackage{subcaption}

\newcommand{\keywords}[1]{%
    \par\noindent\textbf{Keywords:} #1
}

\title{Daily Emergency Call Volume Forecasting using Univariate and Multivariate TimeGPT}

\author[1,2]{Feriel Fass}

\author[1,3]{Djemel Ziou}
\affil[1]{Département d'informatique, Université de Sherbrooke,
Sherbrooke, QC, Canada, J1K 2R1}
\affil[2]{\texttt{Feriel.Fass@usherbrooke.ca}}
\affil[3]{\texttt{Djemel.Ziou@usherbrooke.ca}}

\date{}
\begin{document}

\maketitle
\begin{abstract}
This study investigates the potential of TimeGPT, a pretrained foundation model for time-series, for forecasting daily emergency call volumes. 
Real-world emergency call data, comprising 2,101 daily observations, are used to evaluate both univariate and multivariate forecasting settings under one-step-ahead and seven-day multi-step-ahead scenarios. The multivariate forecasting incorporates temporal, meteorological, demographic, and socioeconomic variables.
A chronological training, validation, and test split is adopted to prevent information leakage, while hyperparameters are selected using the validation set through grid search and early stopping when applicable. Forecasting performance is evaluated using multiple error metrics. Its forecasting performance is compared with several established forecasting approaches, including Seasonal Autoregressive Integrated Moving Average with Exogenous Regressors (SARIMAX), Poisson regression (PR), Negative Binomial regression (NBR), Extreme Gradient Boosting (XGBoost), Multilayer Perceptron (MLP), and Long Short-Term Memory (LSTM). The results show that SARIMAX achieves the best overall performance in the univariate setting. In the multivariate setting, NBR and PR provide best performance. These findings indicate that the large-scale pretraining of TimeGPT do not necessarily translate into superior forecasting accuracy for emergency call volumes. Frugal methods trained solely on the available data remain better for this application.
\end{abstract}


\section{Introduction}
Prehospital emergency services (PES) constitute a critical component of healthcare systems, providing rapid medical assistance to populations in emergency situations. In Quebec, the emergency response process begins when an individual contacts a designated dispatch center to request ambulance services. The call is evaluated by medical dispatchers who determine the appropriate response, which may include dispatching first responders and assigning an ambulance. Paramedics then assess the patient on-site, administer first aid, and if necessary transport them to a healthcare facility or another destination. Over the past decade, PES call volume has increased substantially worldwide \cite{ccsll,fass2025survey}. This sustained growth has been accompanied by longer response times, highlighting increasing challenges in maintaining acceptable service levels \cite{ccsll}. 

These trends emphasize the necessity of improving call volume anticipation to support effective resource planning and operational decision-making. Accurate forecasts of call volumes 
are essential for optimizing the allocation of human and material resources, ensuring acceptable response times \cite{fass2025state,fass2025survey,fildes2002telecommunications,ibrahim2016modeling}. 
Various methods for PES call volume forecasting have been explored in the literature and can be categorized according to modeling approach, and forecast granularity. Modeling approaches mainly include regression-based, autoregressif, neural network-based, and non-parametric methods \cite{fass2025survey, fass2025state}. Forecasting methods can be purely temporal, which capture only time-dependent patterns, or spatiotemporal, which explicitly incorporate spatial information to improve forecasting performance.
With respect to temporal granularity, forecasts are commonly produced at annual, daily, or hourly intervals.
To further enhance forecasting accuracy, many methods integrate explanatory variables reflecting different dimensions of PES demand, including temporal, meteorological, socioeconomic, and demographic factors \cite{fass2025state}.

For daily forecasts, these methods have been evaluated using different performance metrics, including MAPE, and RMSE \cite{fass2025state}. A comparative evaluation of SARIMAX, PR, MLP, LSTM, XGBoost, random forest (RF), gradient boosting regression trees (GBRT), and LightGBM (LGBM) is presented in \cite{fass2025state}. The models incorporate meteorological variables, holidays and weekends, temporal variables, and lagged variables. SARIMAX achieved the smallest forecast error with a MAPE of 6.05\%, and XGBoost the worst with a MAPE of 38.14\%. Despite these advances, emergency call forecasting remains an important area for further improvement.
More recently, foundation models have introduced as a new paradigm for time-series forecasting. Rather than training a model exclusively on the target dataset, these models are pretrained on large collections of heterogeneous time series and are designed to transfer the temporal representations learned during pretraining to new forecasting tasks \cite{benamirouche2026,garza2024timegpt1}. As an implementation of a such paradigm, TimeGPT has been applied to forecasting problems in various domains, such as electricity demand \cite{karim2025multivariate}, and hospital arrivals \cite{zhang2025benchmarking}. However, it seems that it has not yet been used for emergency call volume forecasting. In this work, we investigate the use of univariate and multivariate TimeGPT for emergency call volume forecasting. Its forecasting performance is compared with a broad range of forecasting approaches, including machine-learning methods, namely SARIMAX, PR, and NBR, and XGBoost; and deep-learning methods, namely MLP and LSTM. All methods are evaluated under a common experimental framework to provide a consistent comparison of their forecasting performance.

The rest of this paper is structured as follows. Section \ref{sec2} is devoted to describe the used data. Section \ref{sec3} details the used forecasting methods. Section \ref{sec4} presents the tests conducted and the results reported. Finally, Section \ref{sec5} concludes with a summary of findings and research perspectives.

\section{Data}
\label{sec2}
In this study, experiments are conducted using real emergency call data provided by the health communication center of the Laurentides and Lanaudière regions in Quebec (CCSLL). The dataset contains approximately 600{,}000 calls collected between 2012 and 2017. Each record includes rich temporal, spatial, and contextual attributes, including call timestamps, dispatch and assignment information, geographic coordinates, priority levels, and clinical codes.

To investigate meteorological influences on emergency call, we also use daily weather observations over the same period. The meteorological variables considered in this work include temperature (minimum, maximum, and mean), precipitation, relative humidity, wind speed and direction, visibility, atmospheric pressure, dew point temperature, total rainfall, total snowfall, snow on the ground, and maximum wind gust speed.

Additional socioeconomic and demographic variables, measured on a yearly basis, include total population, the proportions of the population in different age groups (0--4, 5--14, 15--24, 25--34, 35--49, 50--64, and 65 years and older), activity rate, birth, death, and marriage rates, migration-related indicators, family structure, low-income rates, median income, educational attainment, unemployment rate, and the proportion of part-time employment.

For daily granularity, call records are aggregated into non-overlapping 24-hour intervals. 
The resulting daily time series contains 2,101 observations, with daily call volumes ranging from 179 to 469 calls and an average of $\mu=273$ calls/day. 
The variance is $\sigma^2=1395.44$, corresponding to a standard deviation of $\sigma=37.36$. The dispersion factor, computed as $\sigma^2/\mu$, is 5.10, indicating substantial overdispersion in the observed daily call volumes.

A distinction between holidays and non-holidays reveals additional variability in daily emergency call volumes. The dataset contains 76 holiday observations and 2,025 non-holiday observations. During holidays, the average daily call volume increases to 291 calls/day, compared with 272 calls/day on non-holidays, corresponding to an increase of approximately 6.93\%. Moreover, holiday call volumes exhibit substantially greater variability, with a variance of 2788.66 and a standard deviation of 52.81, compared with 1331.55 and 36.49, respectively, for non-holidays. The corresponding dispersion factor increases from 4.88 on non-holidays to 9.55 on holidays, indicating considerably stronger overdispersion during holiday periods.

This distinction aims improve forecasting performance, as suggested in \cite{fass2025state}. The non-holiday dataset was then divided chronologically into training, validation, and test sets. Observations from 2012 to 2015 were used for training, observations from 2016 were used for validation, and observations from 2017 were reserved for testing.

\section{Methods}
\label{sec3}
\begin{figure*}
    \centering
    \includegraphics[width=0.99\linewidth]{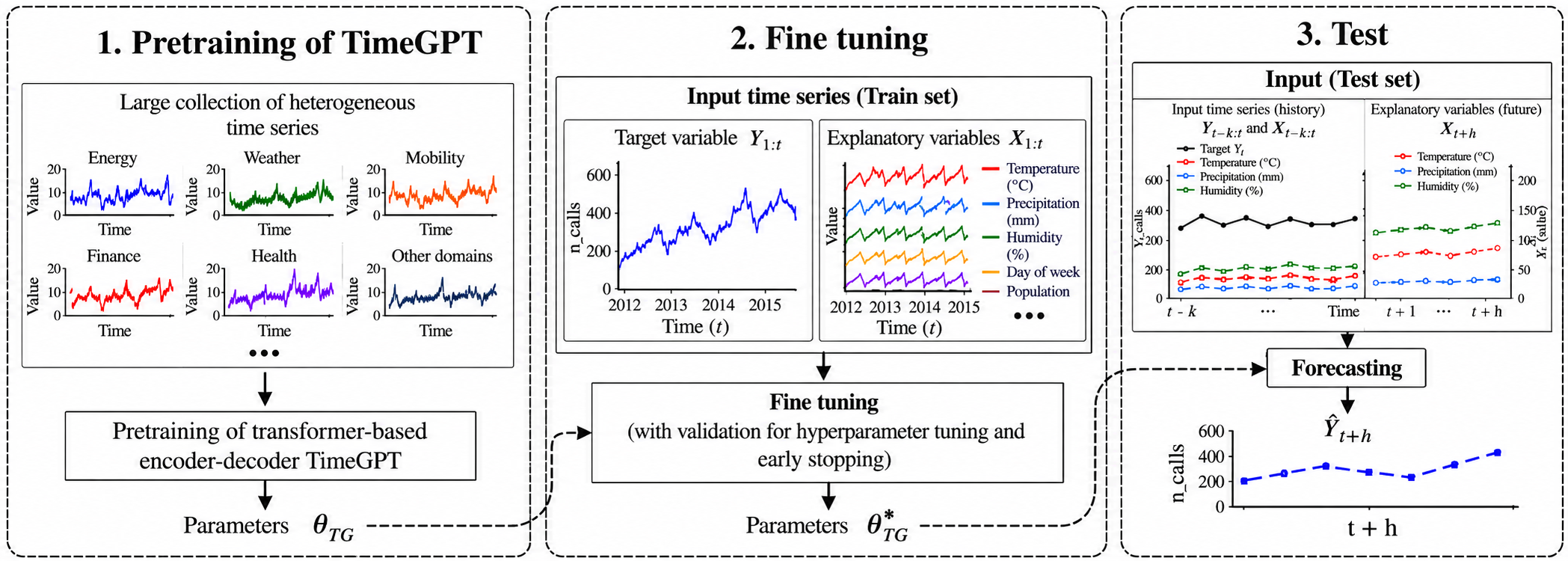}
    \caption{TimeGPT framework for emergency call forecasting, including pretraining, fine-tuning, and testing stages.}
    \label{fig:timegpt}
\end{figure*}

Let \(Y=\{Y_t\}_{t>0}\) denote a discrete-time stochastic process, where \(Y_t\) represents the emergency call volume observed over the interval \([t-g,t]\), and \(g\) is the forecast granularity. Similarly, let \(X=\{X_t\}_{t>0}\) denote a discrete-time stochastic process, where each \(X_t\in\mathbb{R}^{d}\) is a vector of explanatory variables at $t$ defined as
$$
X_t=(X_t^1,X_t^2,\ldots,X_t^d).
$$
These variables capture factors observed over the same interval \([t-g,t]\), including meteorological conditions, temporal characteristics,  demographic, and socioeconomic information associated with emergency calls presented in Section \ref{sec2}. Given the observations available up to time \(t\), \(Y_{1:t}=(Y_1,\ldots,Y_t)\) and \(X_{1:t}=(X_1,\ldots,X_t)\), the objective is to forecast emergency call volumes over a future forecasting horizon \(h\), $
\hat{\mathbf{Y}}_{t+h}.
$
The Forecasting problem is formulated as 
\begin{equation}
\label{eq:F}
\hat{\mathbf{Y}}_{t+h}
=
F\left(
Y_{1:t},
X_{1:t},
X_{t+h};
\theta
\right).
\end{equation}
 Moreover, \(F\) denotes the forecasting method, and \(\theta\) represents its parameters. 

Several methods are considered for specifying \(F\), including TimeGPT, SARIMAX, PR, NBR, RF, GBRT, XGBoost, LGBM, MLP, and LSTM. These methods, except TimeGPT, are described in detail in \cite{fass2025state,fass2025maroc}; therefore, only TimeGPT is introduced in the following.

TimeGPT is a pretrained foundation model designed for time-series forecasting, introduced in \cite{garza2024timegpt1}. Its architecture follows the Transformer paradigm and is based on an encoder--decoder structure adapted to sequential numerical data. Figure~\ref{fig:timegpt} illustrates the three main stages involved in the use of TimeGPT for emergency call forecasting: pretraining, fine-tuning, and testing.

As illustrated in the first block of Figure~\ref{fig:timegpt}, TimeGPT is initially pretrained on a large and heterogeneous collection of time series from multiple domains, such as energy, weather, mobility, finance, and healthcare, comprising approximately 100 billion observation. During this stage, the model learns temporal patterns shared across different time series, such as trends, seasonalities, and short- and long-range dependencies. Let \(\theta_{\mathrm{TG}}\) denote the parameters estimated during this pretraining process. These parameters constitute the preliminary model subsequently transferred to the emergency call forecasting task.

As shown in the second block of Figure~\ref{fig:timegpt}, the estimated parameters \(\theta_{\mathrm{TG}}\) are fine-tuned on the emergency call training set to adapt the model to the characteristics of the target forecasting task giving as output \(\theta_{\mathrm{TG}}^{*}\). The training input consists of the historical emergency call volumes \(Y_{1:t}\) and the corresponding historical explanatory variables \(X_{1:t}\). A validation subset is used for hyperparameter selection and early stopping during the fine-tuning process. Once the hyperparameters estimated, the parameters \(\theta_{\mathrm{TG}}^{*}\) are used for forecasting $Y_{t+h}$ at $t+h$, as illustrated in the third block of Figure~\ref{fig:timegpt}. The used data are the history of emergency call volumes \(Y_{t-k:t}\), the corresponding historical explanatory vectors \(X_{t-k:t}\), and the explanatory vectors available for the future forecasting time \(X_{t+h}\). 

To summarize, rewriting Eq.~\ref{eq:F} to express TimeGPT based forecasting
\begin{equation}
\label{eq:F_TG}
\hat{Y}_{t+h}
=
F_{\mathrm{TG}}
\left(
Y_{t-k:t},
X_{t-k:t},
X_{t+h};
\theta_{\mathrm{TG}}^{*}
\right),
\end{equation}
with
\begin{equation}
\theta_{\mathrm{TG}}
\xrightarrow[\left(Y_{1:t},X_{1:t}\right)]
{\text{fine-tuning}}
\theta_{\mathrm{TG}}^{*}.
\end{equation}

Univariate modeling can be considered a particular case of the multivariate
setting in which forecasting relies exclusively on the historical values of
the target variable, without incorporating explanatory variables. In this case, Eq.~\ref{eq:F_TG} reduces to: 
\begin{equation}
\hat{Y}_{t+h}
=
F_{\mathrm{TG}}
\left(
Y_{t-k:t};
\theta_{\mathrm{TG}}^{*}
\right),
\end{equation}
with 
\begin{equation}
\theta_{\mathrm{TG}}
\xrightarrow[\text{training data }Y_{1:t}]
{\text{fine-tuning}}
\theta_{\mathrm{TG}}^{*}.
\end{equation}

The main distinction between TimeGPT and the other methods considered in this study lies in their learning paradigms. While in the cases of SARIMAX, PR, NBR, XGBoost, MLP, and LSTM the parameters are estimated in one step from training data, those of TimeGPT are estimated in two steps that are pretraining and fine tuning.

The considered forecasting methods also differ substantially in their parameterization. Table~\ref{tab:model_parameters} provides a conceptual comparison of their number of trainable parameters. For all models is considered that the \(d\) denotes the number of explanatory variables, including historical target values gived as a input and a dimension of output is 1.  For PR and NBR, one is added to intercept fit. The additional parameter in NBR corresponds to the dispersion parameter. For SARIMAX, \(p\) and \(q\) denote the non-seasonal autoregressive and moving-average orders, respectively, while \(P\) and \(Q\) denote their seasonal counterparts. For XGBoost, \(M\) denotes the number of trees and \(N_m\) the number of nodes in the \(m\)-th tree.

For the MLP, \(n\) the number of hidden layers, and \(h_k\) the number of units in the \(k\)-th hidden layer. Its parameter count includes the weights connecting successive layers, and the corresponding bias parameters. For the LSTM, \(u\) the number of LSTM units. The factor four in the LSTM parameter count arises from the four sets of parameters associated with the input, forget, output, and candidate-cell mechanisms. For relatively simple architectures, XGBoost, MLP, and LSTM models may involve on the order of thousands of trainable parameters, while more complex configurations can reach millions of parameters. 

Unlike the other models, the exact number of parameters of TimeGPT is not publicly disclosed. However, recent work reports that TimeGPT contains hundreds million parameters \cite{oublal2026timesae}.

Therefore, SARIMAX, PR, NBR are frugal methods, while other methods as TimeGPT are energyvore\footnote{A neologism combining \textit{energy} and \textit{voracious}, used here to denote highly energy-intensive computational models.}. They are computation and memory demanding, as they rely on billions of time-series observations and hundreds of millions of parameters.

\begin{table}[htbp]
\centering
\caption{Conceptual comparison of the number of model parameters.}
\label{tab:model_parameters}
\footnotesize

\begin{tabular}{ll}
\hline
\textbf{Method} & 
\textbf{Approximate number of trainable parameters} \\
\hline

PR &
$d+1$ \\

NBR &
$d+2$ \\

SARIMAX &
$p+q+P+Q+d+1$ \\

XGBoost &
$O\!\left(\sum_{m=1}^{M} N_m\right)$ \\

MLP &
$ 
(d+1)h_1
+
\sum_{k=1}^{n-1}\left(h_k h_{k+1}\right)
+
h_n o
+
\sum_{k=1}^{n} h_k
+
o
$ \\

LSTM &
$\displaystyle
4u(d+u+2)$ \\

TimeGPT &
Not publicly disclosed \\

\hline
\end{tabular}
\end{table}

\section{Tests and results}
\label{sec4}
In this section, we evaluate the performance of the different forecasting methods on the non-holiday dataset presented in Section~\ref{sec2}. As previously described, the dataset was chronologically divided into training, validation, and test sets.

In the following, we first describe the evaluation metrics selected to assess and compare the forecasting performance of the models. We then present the results obtained at a daily temporal granularity for both univariate and multivariate forecasting settings. For each setting, the methods are evaluated under both one-step-ahead and multi-step-ahead forecasting scenarios, and their performance is reported according to the selected evaluation metrics. The methods are also compared against a moving average (MA) baseline, defined as $\hat{Y}_t^{MA} = \frac{1}{k}\sum_{j=0}^{k-1}Y_{t-h-j},$ where (h) denotes the forecasting horizon and (k) is the moving average window size, set to 7 days in this study. The MA forecast corresponds to the arithmetic mean (AM) of the seven most recent daily call volumes available.

Finally, we discuss the obtained results, highlighting the main observations, differences among the methods, and implications of their forecasting performance.

\subsection{Metrics}
To assess models performance, we use several common metrics, including MAE, MAPE, RMSE and RSD. These metrics are standard in call forecasting studies and allow for comparisons with existing research \cite{fass2025state,fass2025maroc,martin2021predicting,steins2019forecasting}. Note that our interpretation of these measures follows the approach to interpreting means presented in \cite{ziou2023pythagorean,ziou2024centrality,ziou2024using}.
The $\text{MAE} = \frac{1}{n} \sum_{i=1}^{n} \left|Y_{i} - \hat{Y}_{i}\right|$ is the AM of the absolute differences between the estimated call volume $\hat{Y}_i$ and the ground truth $Y_i$.
By definition of AM, it is sensitive to outliers. For experiments conducted on the same dataset, smaller MAE values indicate more accurate forecasts.

The $\text{RMSE} = \sqrt{\frac{1}{n} \sum_{i=1}^{n} \left(Y_{i} - \hat{Y}_{i}\right)^2}$ is the square root of the AM of the squared differences between the estimated call volume and the ground truth.
As a mean, the RMSE is always greater than or equal to the MAE, and it is more sensitive than the MAE to outliers. Like MAE, when using the same data, smaller RMSE values indicate a more accurate forecast.

The $\text{MAPE} = \frac{1}{n} \sum_{i=1}^{n} \left| \frac{Y_{i} - \hat{Y}_{i}}{Y_{i}} \right| \times 100$ measures the relative error, expressed as a percentage of the ground truth.
It is scale-free, making it useful for comparing call volume forecasts regardless of the scale. However, it is undefined when the call volume $Y_{i}$ equals zero. Lower MAPE percentages indicating higher accuracy.

The $\text{RSD} = \sqrt{\frac{1}{n} \sum_{i=1}^{n}{\left( (Y_{i} - \hat{Y}_{i})- ME\right)^2}}$, with $ME=\frac{1}{n}\sum_{i=1}^{n}{\left(Y_{i} - \hat{Y}_{i}\right)}$ measures the dispersion of errors around their mean error ($ME$). Lower RSD values indicate that the errors are more closely grouped, reflecting lower dispersion; for example, the largest errors remain close to the $ME$.

The $\mathrm{MASE} =
\frac{\mathrm{MAE}}{\mathrm{MAE}_{MA}} =
\frac{\sum_{i=1}^{n}|Y_i-\hat{Y}_i|}
{\sum_{i=1}^{n}|Y_i-\hat{Y}_i^{MA}|}$,
where $\hat{Y}_i^{MA}$ denotes the forecast obtained using
the MA baseline, measures the forecast error
relative to that of the MA baseline.
Values below, above, or equal to one indicate better, worse, or equivalent performance relative to the MA baseline, respectively.

The minimum and maximum absolute errors, denoted by $\mathrm{Min~AE}= \min_{i=1,\ldots,n}|Y_i-\hat{Y}_i|,$ and $\mathrm{Max~AE}=\max_{i=1,\ldots,n}|Y_i-\hat{Y}_i|$, respectively, characterize the extreme values of the absolute forecast errors. Lower extreme values indicate higher accuracy.

Finally, we report the proportions of forecasts 
$P_{>\tau}=\frac{100}{n} \sum_{i=1}^{n} \mathbb{I}
\left(100\left|\frac{Y_i-\hat{Y}_i}{Y_i}\right|>\tau\right),$ where $\tau\in\{10,20\}$, and $\mathbb{I}(\cdot)$ denotes the indicator function. For the same $\tau$, lower $P_{>\tau}$ values indicate that a smaller proportion of forecasts exceed the corresponding
percentage-error threshold.

\subsection{Hyperparameters tuning}

\begin{table}[ht]
\centering
\caption{Hyperparameter search space considered for the forecasting models.}
\label{tab:hyperparameters}
\scriptsize
\begin{tabular}{p{.8cm}p{3cm}p{3.2cm}}
\hline
\textbf{Model} & \textbf{Hyperparameter} & \textbf{Candidate values} \\
\hline

TimeGPT 
& Fine-tuning steps 
& 0,2,5,10,25,100,500 \\

& Fine-tuning loss 
& default, MAE, MAPE \\

& Fine-tuning depth 
& 1,2,3,4\\

\hline

LSTM 
& Number of LSTM units 
& 32, 64, 128 \\

& Dense units 
& 16, 32 \\

& Dropout 
& 0.0, 0.2, 0.3 \\

& Learning rate 
& $10^{-2}$, $10^{-3}$, $5\times10^{-4}$ \\

& Batch size 
& 32, 64 \\

& Epochs 
& 300 \\
\hline

MLP 
& Hidden-layer units 
& (32,16), (64,32), (128,64) \\

& Dropout 
& 0.0, 0.2, 0.3 \\

& Learning rate 
& $10^{-2}$, $10^{-3}$, $5\times10^{-4}$ \\

& Batch size 
& 32, 64 \\

& Epochs 
& 300 \\
\hline

XGBoost 
& Number of estimators 
& 500, 1000 \\

& Maximum tree depth 
& 3, 5, 7 \\

& Learning rate 
& 0.01, 0.03, 0.05 \\

& Subsample ratio 
& 0.8, 1.0 \\

& Column subsample ratio 
& 0.8, 1.0 \\

& $L_2$ regularization ($\lambda$)
& 1.0, 5.0 \\

& Minimum child weight 
& 1, 5, 10\\
\hline

SARIMAX 
& Non-seasonal order $(p,d,q)$
& (1,0,0), \dots, (2,2,2) \\

& Seasonal order $(P,D,Q,s)$
& (0,0,0,0), (1,0,0,7), \dots, (1,1,1,7) \\

& Trend 
& None, constant, linear \\
\hline

PR 
& Fit intercept 
& False, True \\
\hline

NBR 
& Dispersion parameter $\alpha$ 
& 0.05, 0.1, 0.25, 0.5, 1.0, 2.0 \\

& Fit intercept 
& False, True \\
\hline
\end{tabular}
\end{table}
 
Hyperparameter tuning was performed using a grid-search procedure over a predefined set of values.  Table~\ref{tab:hyperparameters} summarizes the hyperparameters values considered for each method. Each value was evaluated on the validation set using the MAE as a tuning metric, and the value achieving the lowest validation MAE was retained. For models trained iteratively, early stopping was applied by monitoring the validation loss, with a patience of five epochs and a minimum improvement threshold of $10^{-2}$. Training was stopped when the validation loss failed to improve by at least this threshold for five consecutive epochs. The selected hyperparameter configuration was then used to retrain the model on the combined training and validation sets before its final evaluation on the test set. For TimeGPT, the search space included fine-tuning-specific parameters, namely the number of fine-tuning steps, the fine-tuning loss, and the fine-tuning depth.

\begin{table*}[htbp]
\centering
\footnotesize
\caption{Performance comparison of the forecasting models for the univariate one-step-ahead scenario ($h=1$).}
\label{tab:univariate_one_step}
\resizebox{\textwidth}{!}{%
\begin{tabular}{lrrrrrrrrr}
\hline
\textbf{Model} & \textbf{MAE (95\% CI)} & \textbf{MAPE (\%) (95\% CI)} & \textbf{MASE} & \textbf{RMSE} & \textbf{RSD} & \textbf{Min AE} & \textbf{Max AE} & \textbf{$P_{>10\%}$} & \textbf{$P_{>20\%}$} \\
\hline
TimeGPT & 25.380 (23.514, 27.309) & 8.462 (7.869, 9.067) & 1.000 & 31.916 & 31.868 & 0.146 & 90.991 & 32.479 & 7.407 \\
SARIMAX & \textbf{19.281} (17.570, 21.157) & \textbf{6.251} (5.771, 6.771) & \textbf{0.760} & \textbf{24.831} & \textbf{24.353} & 0.018 & \textbf{83.917} & \textbf{18.519} & \textbf{1.425} \\
NBR & 25.803 (23.882, 27.868) & 8.325 (7.771, 8.902) & 1.017 & 32.372 & 31.102 & 0.404 & 109.563 & 34.188 & 3.989 \\
PR & 25.807 (23.884, 27.869) & 8.326 (7.772, 8.903) & 1.017 & 32.377 & 31.105 & 0.293 & 109.567 & 34.188 & 3.989 \\
LSTM & 27.338 (24.741, 30.341) & 8.648 (7.981, 9.395) & 1.078 & 35.181 & 32.075 & 0.007 & 117.110 & 34.758 & 6.268 \\
MLP & 25.632 (23.743, 27.657) & 8.251 (7.731, 8.814) & 1.010 & 32.418 & 31.112 & 0.181 & 103.951 & 31.909 & 5.413 \\
XGBoost & 26.814 (24.750, 28.848) & 8.626 (8.029, 9.192) & 1.057 & 33.916 & 32.712 & 0.052 & 120.239 & 33.618 & 5.413 \\
MA baseline & 25.369 (23.527, 27.296) & 8.409 (7.813, 9.009) & 1.000 & 31.919 & 31.964 & 0.143 & 105.000 & 32.764 & 6.268 \\
\hline
\end{tabular}%
}
\end{table*}
\begin{table*}[htbp]
\centering
\footnotesize
\caption{Performance comparison of the forecasting models for the univariate multi-step-ahead scenario ($h=7$). }
\label{tab:univariate_multi_step}
\resizebox{\textwidth}{!}{%
\begin{tabular}{lrrrrrrrrr}
\hline
\textbf{Model} & \textbf{MAE (95\% CI)} & \textbf{MAPE (\%) (95\% CI)} & \textbf{MASE} & \textbf{RMSE} & \textbf{RSD} & \textbf{Min AE} & \textbf{Max AE} & \textbf{$P_{>10\%}$} & \textbf{$P_{>20\%}$} \\
\hline
TimeGPT & 25.730 (23.619, 27.863) & 8.562 (7.912, 9.243) & 0.969 & 32.504 & 32.481 & 0.308 & 101.898 & 31.909 & 7.407 \\
SARIMAX & \textbf{22.602} (20.950, 24.314) & \textbf{7.473} (6.960, 7.994) & \textbf{0.851} & \textbf{28.888} & \textbf{28.926} & 0.059 & \textbf{94.883} & \textbf{26.211} & \textbf{4.843} \\
NBR & 27.811 (25.272, 30.518) & 8.827 (8.159, 9.546) & 1.047 & 35.354 & 32.120 & 0.323 & 115.156 & 39.316 & 6.838 \\
PR & 27.813 (25.273, 30.520) & 8.827 (8.160, 9.546) & 1.047 & 35.356 & 32.121 & 0.293 & 115.143 & 39.316 & 6.838 \\
LSTM & 26.943 (25.013, 28.986) & 9.127 (8.421, 9.842) & 1.014 & 33.664 & 32.493 & 0.008 & 100.090 & 35.328 & 8.547 \\
MLP & 27.733 (25.214, 30.383) & 8.809 (8.172, 9.499) & 1.044 & 35.256 & 32.368 & 0.128 & 109.297 & 35.897 & 6.268 \\
XGBoost & 29.739 (26.895, 32.914) & 9.333 (8.604, 10.120) & 1.120 & 37.892 & 32.901 & 0.107 & 127.057 & 40.171 & 7.123 \\
MA baseline & 26.564 (24.295, 28.821) & 8.737 (8.062, 9.380) & 1.000 & 33.788 & 33.829 & \textbf{0.000} & 111.000 & 36.182 & 7.407 \\
\hline
\end{tabular}%
}
\end{table*}
\subsection{Univariate Forecast}
First, we present the results obtained using the univariate models, where forecasts are generated using only the number of emergency calls. Lagged target values were constructed using lags of 1, 2, 3, 7, and 14 days. For univariate forecasting, both one-step-ahead and multi-step-ahead forecasting scenarios are evaluated, corresponding to the forecasting horizon, denoted by $h$ in Equations~\ref{eq:F} and~\ref{eq:F_TG}. In the one-step-ahead setting, the forecasting horizon is set to $h=1$, whereas in the multi-step-ahead setting, a horizon of $h=7$ days is considered.

Table~\ref{tab:univariate_one_step} summarizes the forecasting performance of the different models when ($h=1$). The scores indicate that SARIMAX achieves the best overall forecasting performance, outperforming all other models in terms of MAE, RMSE, and MAPE, with values of 19.281, 24.831, and 6.251\%, respectively. SARIMAX also exhibits relatively low residual variability, with a RSD of 24.353. Among the remaining models, MLP achieves the second-best performance according to the MAPE metric, followed by NBR, PR, and MA. TimeGPT ranks behind these models, with its MAPE being approximately 35.37\% higher than that of SARIMAX. Nevertheless, in terms of MAPE, it outperforms LSTM and XGBoost by approximately 2.15\% and 1.90\%, respectively. In terms of MASE, only SARIMAX outperforms the MA baseline, while all other models exhibit comparable or worse performance.

SARIMAX records the smallest Max AE among all evaluated models, at approximately 84 calls, followed by TimeGPT at approximately 91 calls and MLP at 104 calls, while the Max AE exceeds 109 calls for the PR, NBR, LSTM, and XGBoost models. However, the low Max AE of TimeGPT does not necessarily indicate that it is less prone to large relative forecast errors than PR, NBR, MLP, or XGBoost. In fact, $P_{>20}$=7.407\% of TimeGPT forecasts exhibit a relative error greater than 20\% of the observed call volume, which is approximately 85.68\%, 36.84\%, and 18.17\% higher than the corresponding proportions observed for PR/NBR, XGBoost/MLP, and moving average respectively.

Table~\ref{tab:univariate_multi_step} reports the forecasting performance of the evaluated models under the univariate multi-step-ahead setting ($h=7$). Compared with the one-step-ahead forecasting scenario, the forecasting errors increase for all models.
As in the one-step-ahead scenario, SARIMAX achieves the best overall performance, with the lowest MAE (22.602), RMSE (28.888), MAPE (7.473\%), and Max AE of approximately 95 calls.
A notable difference from the one-step-ahead results is observed for TimeGPT. While TimeGPT is outperformed by PR, NBR, and MA baseline in terms of MAPE in the one-step-ahead setting, it becomes the second-best-performing model for the seven-day forecasting horizon. In terms of MAPE, TimeGPT exhibits a value approximately 14.57\% higher than that of SARIMAX. However, its MAPE is approximately 2.00\%, 2.80\%, 3.00\%, and 8.26\% lower than those of MA baseline, MLP, NBR/PR, and XGBoost, respectively.

The proportion of larger relative errors, \(P_{>\tau}\), provides a more nuanced comparison. Although TimeGPT achieves lower average forecasting errors than MLP, NBR, and PR, these models exhibit lower \(P_{>20}\) values. Compared with TimeGPT, \(P_{>20}\) is approximately 15.38\% lower for MLP and 7.68\% lower for both NBR and PR.

\begin{figure}
    \centering
    \includegraphics[width=0.99\linewidth]{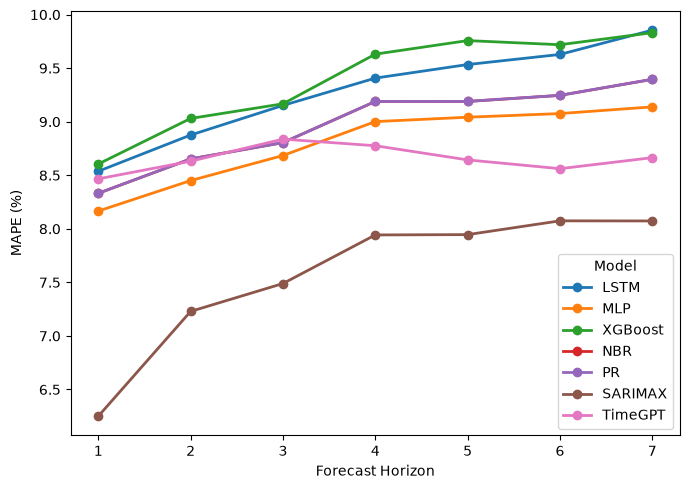}
    \caption{Univariate forecast MAPE evolution within the seven-day horizon.}
    \label{fig:univariate}
\end{figure}

To further analyze forecasting performance over the seven-day horizon, the MAPE is evaluated separately for each forecast day. Figure~\ref{fig:univariate} illustrates the evolution of forecasting MAPE across the seven-day horizon. Overall, most models exhibit a deterioration in forecasting accuracy as the horizon increases, although the magnitude of this deterioration vary considerably across models. SARIMAX consistently achieves the lowest MAPE at every forecast horizon. Its MAPE increases from 6.25\% on the first forecast day to 8.07\% on the seventh day, with most of the increase occurring during the first four horizons, after which its performance remains relatively stable.
TimeGPT also demonstrates relatively stable performance across the seven-day horizon, although it remains less accurate than SARIMAX throughout the forecasting period. Its MAPE increases moderately from 8.47\% at \(h=1\) to a maximum of 8.84\% at \(h=3\), before decreasing to 8.56\% at \(h=6\) and increasing to 8.66\% at \(h=7\). In contrast, LSTM, MLP, XGBoost, PR, and NBR generally show a more pronounced deterioration with increasing horizon. For example, the MAPE of LSTM increases from 8.54\% at \(h=1\) to 9.85\% at \(h=7\), while those of PR and NBR increase from approximately 8.33\% to 9.40\%. XGBoost similarly increases from 8.61\% to 9.83\%, whereas MLP increases from 8.17\% to 9.14\%.
An interesting pattern emerges when comparing TimeGPT with the other methods. At the first forecast horizon, MLP, PR, and NBR achieve lower MAPE values than TimeGPT, while TimeGPT already outperforms LSTM and XGBoost. At the second and third days, MLP remains more accurate than TimeGPT, whereas TimeGPT is close to PR and NBR and more clearly outperforms LSTM and XGBoost. From \(h=4\) onward, however, TimeGPT consistently achieves the second-lowest MAPE after SARIMAX, outperforming LSTM, MLP, XGBoost, PR, and NBR at every remaining forecast horizon. 

\begin{table*}[htbp]
\centering
\footnotesize
\caption{Performance comparison of the forecasting models for the multivariate one-step-ahead scenario ($h=1$). Values in parentheses indicate 95\% confidence intervals. }
\label{tab:multivariate_one_step}
\resizebox{\textwidth}{!}{%
\begin{tabular}{lrrrrrrrrr}
\hline
\textbf{Model} & \textbf{MAE (95\% CI)} & \textbf{MAPE (\%) (95\% CI)} & \textbf{MASE} & \textbf{RMSE} & \textbf{RSD} & \textbf{Min AE} & \textbf{Max AE} & \textbf{$P_{>10\%}$} & \textbf{$P_{>20\%}$} \\
\hline
TimeGPT & 24.726 (22.863, 26.669) & 8.111 (7.524, 8.685) & 0.975 & 31.037 & 30.965 & 0.044 & 94.805 & 33.618 & 4.843 \\
SARIMAX & \textbf{20.307} (18.684, 22.062) & \textbf{6.726} (6.223, 7.270) & \textbf{0.800} & \textbf{25.508} & \textbf{25.535} & 0.114 & \textbf{81.018} & \textbf{24.217} & \textbf{1.994} \\
NBR & 23.146 (21.191, 25.218) & 7.513 (6.973, 8.105) & 0.912 & 30.121 & 29.848 & 0.318 & 119.157 & 28.205 & 3.989 \\
PR & 23.159 (21.214, 25.222) & 7.521 (6.979, 8.106) & 0.913 & 30.122 & 29.865 & 0.239 & 118.446 & 28.490 & 3.989 \\
LSTM & 25.163 (23.086, 27.407) & 8.289 (7.637, 8.927) & 0.992 & 32.311 & 32.336 & 0.046 & 104.507 & 30.199 & 6.553 \\
MLP & 37.078 (33.701, 40.688) & 12.850 (11.607, 14.238) & 1.462 & 44.212 & 33.600 & 0.299 & 119.175 & 53.276 & 18.519 \\
XGBoost & 27.091 (24.917, 29.481) & 8.560 (8.006, 9.132) & 1.068 & 34.743 & 31.325 & 0.114 & 111.914 & 35.328 & 4.843 \\
MA baseline& 25.369 (23.527, 27.296) & 8.409 (7.813, 9.009) & 1.000 & 31.919 & 31.964 & 0.143 & 105.000 & 32.764 & 6.268 \\
\hline
\end{tabular}%
}
\end{table*}
\begin{table*}[htbp]
\centering
\footnotesize
\caption{Performance comparison of the forecasting models for the multivariate multi-step-ahead scenario ($h=7$). Values in parentheses indicate 95\% confidence intervals.}
\label{tab:multivariate_multi_step}
\resizebox{\textwidth}{!}{%
\begin{tabular}{lrrrrrrrrr}
\hline
\textbf{Model} & \textbf{MAE (95\% CI)} & \textbf{MAPE (\%) (95\% CI)} & \textbf{MASE} & \textbf{RMSE} & \textbf{RSD} & \textbf{Min AE} & \textbf{Max AE} & \textbf{$P_{>10\%}$} & \textbf{$P_{>20\%}$} \\
\hline
TimeGPT & 24.930 (23.064, 26.856) & 8.139 (7.618, 8.689) & 0.938 & 31.465 & 31.297 & 0.242 & 107.117 & 34.188 & 5.698 \\
SARIMAX & 24.746 (22.871, 26.890) & 8.202 (7.581, 8.888) & 0.932 & 31.309 & 31.338 & 0.197 & \textbf{104.921} & 32.479 & 5.128 \\
NBR & 23.439 (21.180, 25.898) & 7.568 (6.956, 8.223) & 0.882 & 31.037 & \textbf{30.521} & 0.181 & 129.741 & \textbf{28.490} & \textbf{4.843} \\
PR & \textbf{23.425} (21.184, 25.874) & \textbf{7.567} (6.957, 8.223) & \textbf{0.882} & \textbf{31.012} & 30.524 & 0.268 & 128.978 & \textbf{28.490} & 5.128 \\
LSTM & 43.694 (39.530, 48.106) & 13.559 (12.450, 14.662) & 1.645 & 52.646 & 32.252 & 0.211 & 165.086 & 66.667 & 19.373 \\
MLP & 607.185 (599.292, 614.930) & 201.437 (197.011, 206.048) & 22.857 & 630.168 & 168.875 & 292.804 & 973.597 & 100.000 & 100.000 \\
XGBoost & 28.754 (26.072, 31.695) & 9.033 (8.321, 9.792) & 1.082 & 36.541 & 30.780 & 0.272 & 119.967 & 40.741 & 5.983 \\
MA baseline & 26.564 (24.295, 28.821) & 8.737 (8.062, 9.380) & 1.000 & 33.788 & 33.829 & \textbf{0.000} & 111.000 & 36.182 & 7.407 \\
\hline
\end{tabular}%
}
\end{table*}

\subsection{Multivariate Forecast}
In the multivariate forecasting setting, historical values in addition to variables indicated in Section \ref{sec2} are used. These variables characterize the temporal, meteorological, demographic, and socioeconomic conditions associated with emergency call demand. Table~\ref{tab:multivariate_one_step} reports the forecasting performance obtained by each method under the multivariate one-step-ahead setting ($h=1$). Compared with the univariate setting, all methods achieve a better forecasting performance in the multivariate setting, except MLP and SARIMAX. One possible explanation is that these two methods not adequately capture the relationships between the explanatory variables and emergency call volume.

Unlike in the univariate setting, SARIMAX, TimeGPT, PR, and NBR outperform the MA baseline in terms of MASE. Consistent with the univariate setting, SARIMAX achieves the best overall forecasting performance, exhibiting the lowest errors across all metrics. However, its forecasting accuracy slightly deteriorates after incorporating the explanatory variables. In terms of MAPE, its error is approximately 7.60\% higher than in the univariate setting, suggesting that the relationships between the explanatory variables and emergency call volume is not linear. NBR and PR achieve the second-lowest errors across all metrics, except for Max AE. Compared with SARIMAX, the MAPE of NBR is approximately 11.70\% higher. However, despite their low average forecasting errors, NBR and PR exhibit Max AE values of approximately 119 calls, which are substantially highest Max AE compared to all other methods. TimeGPT ranks behind SARIMAX and PR/NBR according to the main average error metrics. Its MAPE is approximately 7.96\% higher than that of NBR and 20.59\% higher than that of SARIMAX. Nevertheless, it achieves a Max AE of approximately 95 calls, the second-lowest value among all evaluated methods after SARIMAX and considerably lower than those obtained by NBR and PR. TimeGPT also outperforms LSTM, XGBoost, and MLP in terms of MAE, RMSE, and MAPE.

The proportions of relative errors further highlights the differences between the methods. SARIMAX again achieves the best performance, with only $P_{>20\%}$=1.994\% of its forecasts exhibiting relative errors exceeding 20\% of the observed call volume. Consistent with the univariate results, TimeGPT exhibits a lower Max AE than NBR and PR but generates a larger proportion of forecasts with relative errors exceeding 20\%. Specifically, $P_{>20}$=4.843\% for TimeGPT forecasts, which is approximately 21.41\% higher thos of NBR and PR. This proportion further increases for XGBoost, LSTM, and MLP.

Table~\ref{tab:multivariate_multi_step} reports the forecasting performance obtained by each method under the multivariate multi-step-ahead setting ($h=7$). Unlike the one-step-ahead multivariate results, NBR and PR achieve the best overall performance across all metrics except Max AE. In term of MAPE, they achieve 7.567\%.  

TimeGPT and SARIMAX exhibit relatively similar average forecasting errors. TimeGPT achieves a MAPE of 8.139\%, approximately 7.56\% higher than that of PR, while SARIMAX achieves a MAPE of 8.202\%. Thus, TimeGPT outperforms SARIMAX in terms of Max AE, with TimeGPT achieving a Max AE of 107 calls, compared with 116 calls for SARIMAX and approximately 129 calls for both NBR and PR. TimeGPT also outperforms XGBoost, LSTM, and MLP according to all metrics.

The analysis of the relative-error thresholds provides further insight into the distribution of forecasting errors. PR and NBR achieve the lowest proportion of forecasts with relative errors exceeding 10\%, both with \(P_{>10\%}=28.490\%\), approximately 16.67\% lower than TimeGPT. For the 20\% threshold, NBR performs best with \(P_{>20\%}=4.843\%\), followed by PR with \(5.128\%\), corresponding to approximately 15.01\% and 10.00\% lower proportions than TimeGPT, respectively. Thus, despite achieving a lower Max AE than PR and NBR, TimeGPT produces large relative more frequently.

\begin{figure}
    \centering \includegraphics[width=0.99\linewidth]{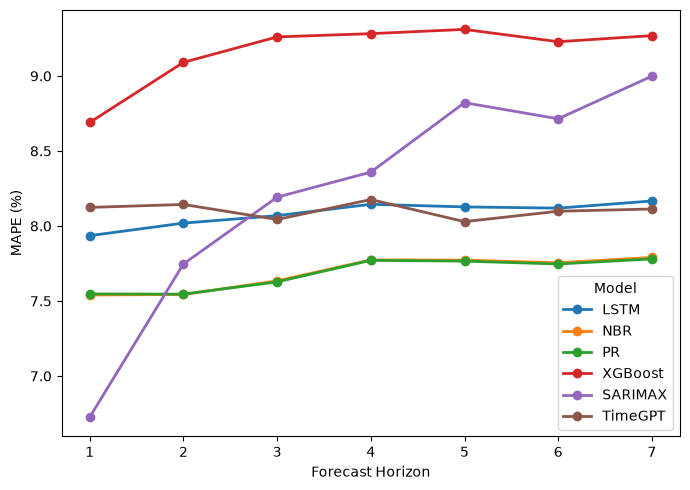}
    \caption{Multivariate forecast MAPE evolution within the seven-day horizon.}
    \label{fig:multivariate}
\end{figure}

To examine how the forecasting error evolves across the seven days of the horizon, the MAPE was calculated separately for each forecast day. Figure~\ref{fig:multivariate} illustrates this evolution for the multivariate methods.
NBR and PR exhibit the best and most stable performance across most of the seven forecast days. Their MAPE values remain very close throughout the entire horizon, starting at approximately 7.54\% for the first forecast day and remaining below 7.8\% across all subsequent forecast days. In contrast, SARIMAX exhibits a more pronounced increase in MAPE across the forecasting horizon, from approximately 6.73\% on the first forecast day to approximately 9.00\% on the seventh forecast day.

TimeGPT exhibits relatively stable performance across the seven-day forecasting horizon, with its MAPE remaining between approximately 8.03\% and 8.18\%. During the first two forecast days, SARIMAX achieves a lower MAPE than TimeGPT. From the third forecast day onward, however, TimeGPT achieves a lower MAPE than SARIMAX. TimeGPT remains less accurate than NBR and PR across the seven forecast days. TimeGPT also consistently outperforms XGBoost in terms of MAPE across the entire seven-day forecasting horizon.

\subsection{Discussion}
\begin{table*}[htbp]
\centering
\footnotesize
\caption{Pairwise Diebold--Mariano test p-values for
univariate and multivariate forecasting settings.
Bold values indicate statistically significant
differences ($p<0.05$).}
\label{tab:dm_comparison}
\begin{tabular}{ccccc}
\hline
\textbf{Forecast} &
\textbf{Horizon} &
\textbf{TimeGPT vs SARIMAX} &
\textbf{TimeGPT vs MA} &
\textbf{SARIMAX vs MA} \\
\hline
Univariate
& $h=1$
& $\mathbf{<0.0001}$
& 0.9817
& $\mathbf{<0.0001}$ \\

& $h=7$
& $\mathbf{<0.0001}$
& 0.2588
& $\mathbf{<0.0001}$ \\
\hline
Multivariate
& $h=1$
& $\mathbf{<0.0001}$
& 0.3881
& $\mathbf{<0.0001}$ \\

& $h=7$
& 0.8460
& 0.0535
& $\mathbf{0.0424}$ \\
\hline
\end{tabular}%
\end{table*}
Across both univariate and multivariate settings, as well as for one-step-ahead and multi-step-ahead forecasting, frugal approaches such as PR, NBR, and SARIMAX achieve better forecasting performance. This result is particularly noteworthy considering that, unlike TimeGPT, these models are fitted only on the available training data, comprising only 1320 observations, and involve approximately 50 parameters. Nevertheless, they are able to effectively capture the underlying patterns of emergency call demand and outperform the pretrained model in several experimental settings.

To further assess whether these differences are statistically
significant, pairwise Diebold--Mariano (DM) tests were
conducted using absolute forecast errors \cite{Diebold01071995}.
Table~\ref{tab:dm_comparison} reports the corresponding
p-values for TimeGPT, SARIMAX, and the MA baseline.
Bold values indicate statistically significant differences
at a threshold of 0.05. The DM tests show that the
differences between SARIMAX and TimeGPT forecasts are
significant in three of the four experimental scenarios.
The only exception is the multivariate multi-step-ahead
setting. More importantly, the differences between
TimeGPT and the MA baseline forecasts are not significant
in any of the four scenarios. In contrast, the differences
between SARIMAX and the MA baseline are significant
across all four scenarios.

These findings suggest that large-scale pretraining on approximately 100 billion observations and the use of hundreds of millions of parameters do not necessarily translate into superior forecasting accuracy for emergency call demand. Instead, frugal methods or a simple MA remain performing than energyvore methods in this application.

Furthermore, comparing the univariate and multivariate forecasting results shows that incorporating the explanatory variables presented in Section \ref{sec2} leads to only limited improvements in forecasting performance for most of the best-performing methods. Several hypotheses may explain this observation. First, the relationships between some explanatory variables and emergency call demand may not be adequately captured by the functional forms assumed by the evaluated methods.

Second, not all explanatory variables may provide relevant or complementary predictive information about emergency call demand. Including weakly informative, redundant, or highly correlated variables may therefore provide little additional predictive benefit. This suggests that an explicit variable-selection or feature-importance analysis could be beneficial to identify the subset of explanatory variables that contributes most to forecasting performance. Methods such as SHAP could be used to assess the contribution of individual features to model forecasting. In addition, causal inference approaches could be investigated to distinguish variables that are merely associated with emergency call demand from those that may have a causal influence on demand.

\section{Conclusion}
\label{sec5}
In this study, we investigated the potential of TimeGPT for forecasting emergency call demand using real-world emergency call data aggregated at a daily temporal granularity. The performance of TimeGPT was compared with several forecasting approaches commonly used in the literature, including frugal methods (SARIMAX, PR, and NBR), machine-learning methods (XGBoost), and deep-learning methods (MLP and LSTM). Both univariate and multivariate forecasting settings were considered, and the methods were evaluated under one-step-ahead and multi-step-ahead forecasting scenarios. Hyperparameter tuning was also performed for the applicable machine- and deep-learning methods to ensure a fair comparison. The experimental results show that, despite being energyvore, TimeGPT can be competitive with some of the machine- and deep-learning methods, but does not outperform frugal methods or moving average baseline considered in this study. In particular, SARIMAX provides the strongest performance in the univariate setting, whereas PR and NBR achieve the best overall performance when explanatory variables are incorporated. Future work should therefore investigate more systematic feature-selection and feature-importance strategies to identify the explanatory variables that contribute most to emergency call forecasting.

\bibliography{sn-bibliography}
\bibliographystyle{ieeetr}

\end{document}